\pdfoutput=1

\documentclass[11pt]{article}

\usepackage[final]{acl}

\usepackage{times}
\usepackage{latexsym}

\usepackage[T1]{fontenc}

\usepackage[utf8]{inputenc}

\usepackage{microtype}

\usepackage{inconsolata}

\usepackage{graphicx}
\usepackage{multirow}
\usepackage{booktabs}   
\usepackage{subcaption} 
\usepackage{rotating}   
\title{Benchmarking and Improving LVLMs on Event Extraction \\ from Multimedia Documents}

\author{Fuyu Xing\thanks{Equal contribution.}, Zimu Wang\footnotemark[1], Wei Wang, Haiyang Zhang\thanks{Corresponding author.} \\
  School of Advanced Technology, Xi'an Jiaotong-Liverpool University \\
  \texttt{\{Fuyu.Xing21,Zimu.Wang19\}@student.xjtlu.edu.cn} \\
  \texttt{\{Wei.Wang03,Haiyang.Zhang\}@xjtlu.edu.cn} \\}

\begin{document}
\maketitle
\begin{abstract}
The proliferation of multimedia content necessitates the development of effective Multimedia Event Extraction (M²E²) systems. Though Large Vision-Language Models (LVLMs) have shown strong cross-modal capabilities, their utility in the M²E² task remains underexplored. In this paper, we present the first systematic evaluation of representative LVLMs, including DeepSeek-VL2 and the Qwen-VL series, on the M²E² dataset. Our evaluations cover text-only, image-only, and cross-media subtasks, assessed under both few-shot prompting and fine-tuning settings. Our key findings highlight the following valuable insights: (1) Few-shot LVLMs perform notably better on visual tasks but struggle significantly with textual tasks; (2) Fine-tuning LVLMs with LoRA substantially enhances model performance; and (3) LVLMs exhibit strong synergy when combining modalities, achieving superior performance in cross-modal settings. We further provide a detailed error analysis to reveal persistent challenges in areas such as semantic precision, localization, and cross-modal grounding, which remain critical obstacles for advancing M²E² capabilities.
\end{abstract}

\section{Introduction}

Event extraction (EE) is a fundamental task in NLP, aiming to identify structured events from unstructured data \cite{Ahn2006,peng2023omnievent}. It typically involves two subtasks: event detection (ED), which detects the event trigger words and classifies their event type, and event argument extraction (EAE), which extracts entities and labels their argument roles. Traditional EE methods have primarily focused on a single modality, either text \cite{wang2022maven,wang2024document,ZhangJi2021}, image \cite{Yatskar2016,Cho2022,Pratt2020}, or video \cite{Soomro2012_UCF101,Ye2015,Heilbron2015}.

\begin{figure}[t]
    \centering
    \includegraphics[width=\columnwidth]{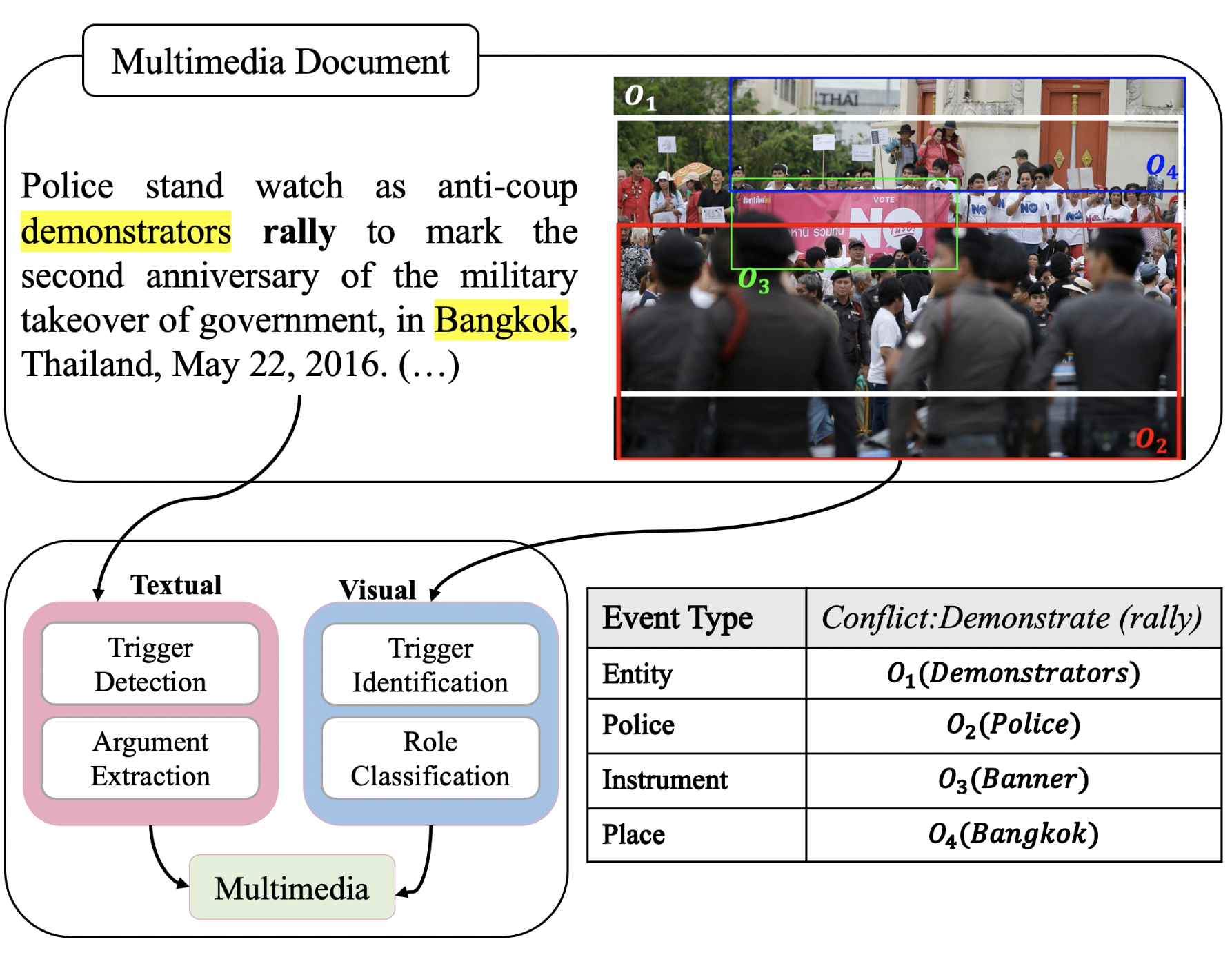}
    \caption{Illustration of the M²E² task, including an image and accompanying text, where event arguments may appear in either or both modalities.}
    \vspace{-4mm}
    \label{fig1:m2e2}
\end{figure}

In the era of digital media, content like news articles often contains both textual and visual modalities, and some arguments presented in images may not appear in the accompanying text. This has prompted the rise of Multimedia Event Extraction (M²E², \citealp{Li2020_CrossMedia}), targeting to extract information from combined texts and images (see Figure \ref{fig1:m2e2}). Prior works either rely on limited hand-crafted features or require expensive annotation efforts. Meanwhile, Large Vision-Language Models (LVLMs) have demonstrated strong cross-modal reasoning abilities \cite{widm.1574,11033252}, but their utility for M²E² remains underexplored.

Motivated by this gap, we present the first systematic evaluation of LVLMs on the M²E² dataset. Evaluations are conducted across six subtasks for ED and EAE in text-only, image-only, and cross-modal settings, under both few-shot and fine-tuning paradigms. Our key findings reveal that: (1) Few-shot LVLMs demonstrate strong performance compared to specialized baselines in visual-only EE but lag behind established pre-trained language models (PLM) baselines textually. (2) Fine-tuning LVLMs dramatically improves their capabilities, enabling smaller models to surpass strong textual baselines and achieve state-of-the-art results. (3) Utilizing LVLMs on multimedia EE consistently yields the best results, showcasing the effective cross-modal synergy of LVLMs. (4) Despite performance improvements, persistent challenges exist across all modalities, including difficulties with semantic discrimination, localization, effective cross-modal grounding, and safety filter behaviors.

\section{Related Work}

\paragraph{Event Extraction} has mainly been investigated in single modalities. Textual EE has progressed from token-level classification \cite{Ramponi2020,Yang2024_MLEE} to generative models \cite{peng2023does,qi2024adelie}.
On the other hand, visual EE has traditionally focused on coarse ED from images or videos \cite{Gu2018_Ava,Wu2019_LongTerm}, relying on predefined event types or bounding boxes \cite{Yatskar2016,SilbererPinkal2018}. Recent methods remove such constraints by integrating object detection and attention-based reasoning \cite{Li2020_CrossMedia}.
Emerging studies have also highlighted the improvements by integrating external visual knowledge or multimedia parameters on textual EE, showcasing the complementary strengths of different modalities \cite{JiGrishman2008,Zhang2017_Multimodal,ZhangJi2018_GAIL}.

\paragraph{Multimedia Event Extraction} combines textual and visual modalities to improve robustness, employing fusion strategies, such as feature-level concatenation \cite{Horng2017,Chen2018_Postop}, shared latent spaces \cite{Khadanga2019}, and decision-level aggregation. Recent frameworks such as WASE \cite{Li2020_CrossMedia}, CLIP-Event \cite{Li2022_CLIPEvent}, and CAMEL \cite{Du2023_Generated} incorporate weak supervision or inject structured event schemas to improve multimodal grounding. Instruction-tuned models like UMIE \cite{Sun2024_UMIE} and video-text paired learning \cite{cao2025cross} enhance event representation across modalities. However, despite the success of generative models in event understanding and reasoning tasks \cite{li2024meqa,li2025efficient}, their potential in the critical M²E² remains underexplored. We pioneer this field by utilizing LVLMs to M²E², offering valuable analysis and insights for future research. 


\section{Problem Formulation}

\begin{figure*}[t!]
    \centering
    \includegraphics[width=\textwidth]{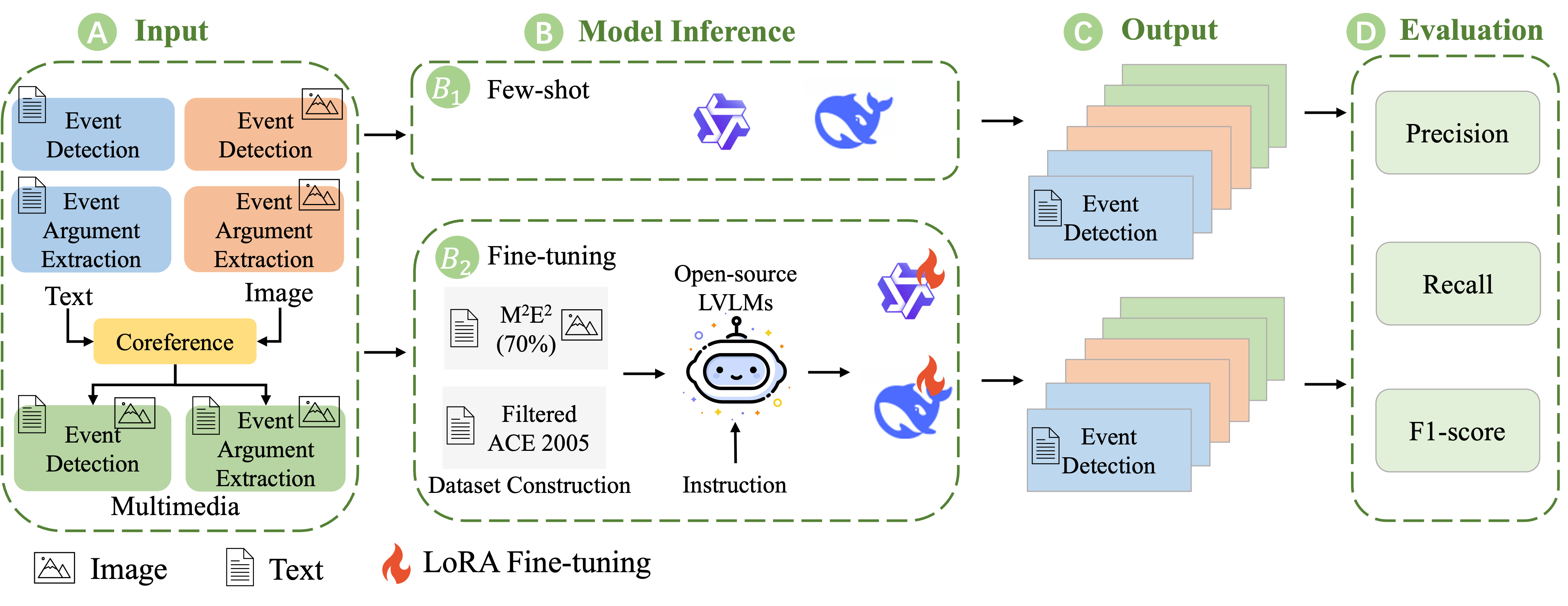}
    \vspace{-6mm}
    \caption{Evaluation pipeline under few-shot and fine-tuning paradigms. A multimedia document is processed, and the extracted events and arguments are compared against ground-truth annotations with strict matching.
    }
    \label{fig:eval_flowchart}
    \vspace{-4mm}
\end{figure*}

We focus on the textual, visual, and multimedia EE defined by \newcite{Li2020_CrossMedia}. Each \textit{Event} includes a trigger and one or more arguments. Formally:
(1) \textit{Event Trigger} (\(\tau\)) is a verb or noun in the text indicating the occurrence of the event.
(2) \textit{Event Argument} (\(\alpha\)) plays a role in the event, either a text entity, a temporal expression, a numerical value, or a detected object in the image.
(3) \textit{Argument Role} (\(\rho\)) is the semantic relationship linking an argument \(\alpha\) to its event mention. Our two main tasks over a multimodal document are as follows:

\paragraph{Event Detection} seeks to detect all event mentions \{\(EM_1\), …, \(EM_n\)\} within the input context. Each \(EM_i\) is assigned a category \(c_i\in C\) and is grounded in one or both modalities:
\[EM_i = (c_i, G_i), \ G_i \subseteq \{\tau_i, v_i\},\]
where \(\tau_i\) is the trigger and \(\upsilon_i\) is the corresponding image. 
Based on the content of the grounding set \(G_i\), an event mention can be classified into one of the following types: \textbf{text-only} event mention where \(G_i=\{\tau_i \}\), \textbf{image-only} event mention where \( G_i=\{\upsilon_i\}\)  and \textbf{multimodal} event mention where \(G_i=\{\tau_i, \upsilon_i\}\), indicating that the same event is referred to both textually and visually.

\paragraph{Event Argument Extraction} extracts a set of arguments \(\{\alpha_{i1},…,\alpha_{ij}\}\) for each identified \(EM_i\). Each argument \(\alpha_{ij}\) is represented as:
\[\alpha_{ij} = (\rho_{ij}, H_{ij}), \ H_{ij} \subseteq \{e_{ij}, o_{ij}\},\]
where \(\rho_{ij}\in R\) is the argument role and \(H_{ij} \subseteq \{e_{ij}, o_{ij}\}\) indicates grounding in textual entity \(e_{ij}\) or visual object \(o_{ij}\) or both. Arguments referring to the same real-world object in text and image are merged into a single unified representation.

\section{Evaluation Settings}

\subsection{Datasets and Evaluation Metrics}
We conduct our experiments on M²E² \cite{Li2020_CrossMedia}, a cross-media EE dataset of news articles, containing $1,105$ text-only mentions, $188$ image-only mentions, and $395$ cross-media mentions. We utilize $70\%$ for training and $30\%$ for evaluation. All events are categorized into $8$ fine-grained subtypes (e.g., \texttt{Movement.Transport}) with $18$ argument roles. For auxiliary data, we leverage ACE 2005 \cite{Walker2006_ACE2005}, filtering it to match the eight event types in M²E². A conceptual overview of our evaluation pipeline is presented in Figure \ref{fig:eval_flowchart}.

For all subtasks, we report the precision, recall, and F1-score under the following conditions:
(1) \textbf{Text-only ED}, if both the trigger word and its event type are matched;
(2) \textbf{Image-only ED}, if the image and event type are correct;
(3) \textbf{Text-only EAE}, if both the argument and its role are matched;
(4) \textbf{Image-only EAE}, if the argument and role match and the Intersection over Union (IoU) between the predicted and ground truth bounding box exceeds $0.5$;
(5) \textbf{Cross-media ED}, if the trigger (or image) and event type match exactly; and
(6) \textbf{Cross-media EAE}, if the argument (or bounding box) and role are matched.


\subsection{Prompt Design}
Table \ref{tab:prompt_example} shows an example of the few-shot prompt used to guide the LVLMs. Each prompt consists of the following key components: a task-specific \textbf{Instruction}, four contrastive \textbf{Demonstrations} (both positive and negative) to enhance task understanding, the \textbf{Query} input (text and/or image), a set of candidate \textbf{Options} (e.g., event or role types), and a final \textbf{Answer} placeholder for the model's output. For the EAE subtask, the ground-truth \textbf{Event Type} is also supplied as context. In fine-tuning experiments, the structure of the prompt is similar, while the demonstrations are removed.

\subsection{Experimental Setup}
We conduct experiments across a range of open-source LVLMs: DeepSeek-VL2 \cite{wu2024deepseekvl2}, Qwen2-VL-7B \cite{bai2025qwen25vl}, Qwen2.5-VL-7B, and Qwen2.5-VL-72B \cite{bai2025qwen25vl}. Our evaluation includes a \textit{few-shot} paradigm to assess out-of-the-box capabilities and a \textit{fine-tuning} paradigm to measure performance after adaptation.
During the fine-tuning process, we adapt Qwen2-VL-7B and Qwen2.5-VL-7B using LoRA \cite{Hu2022_LoRA}.
We train these models on a combined dataset created by merging the M²E² training split with filtered ACE 2005, separately for each sub-task. The fine-tuning is carried out with a learning rate of \(1\times10^{-4}\) and batch size of $3$. We keep the inference temperature at $0$ to ensure output stability. All experiments are conducted on $2$ NVIDIA GeForce RTX 4090 graphics cards.

\subsection{Baselines}
We compare the performance of LVLMs against the following baselines across all modalities. These include unimodal methods such as the graph-based \textbf{JMEE} \cite{Liu2018_JointMulti} for text and \textbf{Clip-Event} \cite{Li2022_CLIPEvent} for vision. For multimedia models, we compare against diverse strategies, including those based on shared semantic embeddings (\textbf{WASE}, \citealp{Li2020_CrossMedia}, \textbf{UniCL}, \citealp{Liu2022_UniCL}), data synthesis (\textbf{CAMEL}, \citealp{Du2023_Generated}), instruction-tuning (\textbf{UMIE}, \citealp{Sun2024_UMIE}), and the current state-of-the-art multi-task framework, \textbf{X-MTL} \cite{cao2025cross}.

\subsection{Experimental Results}

\begin{table*}[t!]
\centering
\small
\setlength{\tabcolsep}{3pt} 

\begin{tabular}{ll|c|cccc|cc}
\toprule
\multirow{2}{*}{\textbf{Task}} & \multirow{2}{*}{\textbf{Metric}} & \textbf{SOTA} & \multicolumn{4}{c|}{\textbf{Few-shot LVLMs}} & \multicolumn{2}{c}{\textbf{Fine-tuned (LoRA)}} \\
\cmidrule(lr){3-3} \cmidrule(lr){4-7} \cmidrule(lr){8-9}
 & & \textbf{X-MTL} & \textbf{DS-VL} & \textbf{Qwen2-VL} & \textbf{Qwen2.5-VL} & \textbf{Qwen2.5-VL-72B} & \textbf{Qwen2-VL} & \textbf{Qwen2.5-VL} \\
\midrule
\multicolumn{9}{c}{\textbf{Textual Event Extraction}} \\
\midrule
\multirow{3}{*}{Event Mention} 
 & P & $49.7$ & $30.5$ & $13.3$ & $70.0$ & $80.0$ & $46.8$ & $41.5$ \\
 & R & $65.7$ & $12.0$ & $19.7$ & $3.7$ & $12.0$ & $59.6$ & $52.6$ \\
 & F1 & $\mathbf{56.6}$ & $17.5$ & $15.9$ & $6.9$ & $20.6$ & $52.5$ & $46.4$ \\
\cmidrule(lr){2-9}
\multirow{3}{*}{Argument Role} 
 & P & $34.6$ & $5.6$ & $75.0$ & $4.2$ & $10.1$ & $16.3$ & $35.5$ \\
 & R & $37.6$ & $2.2$ & $2.6$ & $1.4$ & $7.5$ & $24.1$ & $50.3$ \\
 & F1 & $36.0$ & $3.1$ & $5.1$ & $2.1$ & $8.6$ & $19.5$ & $\mathbf{41.6}$ \\
\midrule
\multicolumn{9}{c}{\textbf{Visual Event Extraction}} \\
\midrule
\multirow{3}{*}{Event Mention}
 & P & $73.1$ & $39.3$ & $69.6$ & $54.3$ & $73.0$ & $49.7$ & $48.5$ \\
 & R & $70.3$ & $76.4$ & $62.1$ & $47.3$ & $70.6$ & $74.4$ & $72.5$ \\
 & F1 & $71.7$ & $51.9$ & $65.6$ & $50.5$ & $\mathbf{71.8}$ & $59.6$ & $58.1$ \\
\cmidrule(lr){2-9}
\multirow{3}{*}{Argument Role} 
 & P & $33.2$ & $0.7$ & $3.3$ & $23.7$ & $62.8$ & $23.5$ & $10.8$ \\
 & R & $31.3$ & $0.7$ & $4.3$ & $35.3$ & $71.9$ & $4.3$ & $19.6$ \\
 & F1 & $32.2$ & $0.7$ & $3.8$ & $28.3$ & $\mathbf{67.0}$ & $7.3$ & $14.0$ \\
\midrule
\multicolumn{9}{c}{\textbf{Multimedia Event Extraction}} \\
\midrule
\multirow{3}{*}{Event Mention} 
 & P & $78.3$ & $41.3$ & $65.4$ & $62.9$ & $84.0$ & $41.7$ & $56.6$ \\
 & R & $57.3$ & $79.6$ & $53.7$ & $37.2$ & $81.4$ & $79.3$ & $77.4$ \\
 & F1 & $66.2$ & $54.4$ & $60.0$ & $46.8$ & $\mathbf{82.5}$ & $54.7$ & $65.4$ \\
\cmidrule(lr){2-9}
\multirow{3}{*}{Argument Role}
 & P & $40.3$ & $8.1$ & $20.6$ & $23.9$ & $56.1$ & $55.3$ & $54.7$ \\
 & R & $42.6$ & $5.8$ & $21.9$ & $14.9$ & $71.9$ & $70.8$ & $80.1$ \\
 & F1 & $41.4$ & $6.8$ & $21.2$ & $18.4$ & $63.0$ & $62.1$ & $\mathbf{65.0}$ \\
\bottomrule
\end{tabular}
\caption{Main results on the M²E² benchmark. \textbf{Bold} indicates the best F1-score in each modality for each task. (DS-VL: DeepSeek-VL2, Qwen2-VL: Qwen2-VL-7B, Qwen2.5-VL: Qwen2.5-VL-7B)}
\vspace{-4mm}
\label{tab:results}
\end{table*}

Table \ref{tab:full_results} presents the evaluation results of LVLMs in comparison to baseline models, where X-TML, the state-of-the-art multi-task framework, is included. The full, unabridged table featuring all baseline models is provided in Appendix \ref{sec:full}. From the table, we have the following observations:

First, a notable performance disparity exists in the textual domain, where few-shot LVLMs fall significantly short compared to specialized models such as X-MTL. However, fine-tuning LVLMs effectively bridges this gap. For instance, Qwen2-VL-7B with LoRA yields a substantial $36.6\%$ increase in F1-score for ED and a $14.4\%$ increase in EAE, elevating its performance to be on par with strong textual baselines. Notably, the fine-tuned Qwen2.5-VL-7B surpassed these baselines in EAE, highlighting that even smaller, fine-tuned LVLMs are capable of achieving state-of-the-art results on complex textual tasks.

Second, in the visual domain, the performance trend is reversed. Few-shot LVLMs, particularly Qwen2.5-VL-72B, exhibit exceptional out-of-the-box capabilities, surpassing all specialized visual baselines in both ED (achieving an F1-score of $71.8\%$) and EAE (with an F1-score of $67.0\%$). This indicates that their pre-training effectively captures the knowledge required for visual event understanding, though argument localization remains open challenges for models that have not been explicitly pre-trained on real-world coordinates.

Finally, LVLMs consistently achieved the highest overall performance in the multimedia setting, underscoring the complementary relationship between textual and visual modalities. Notably, even models that struggled with text-only ED show significant improvements when paired with the corresponding images, highlighting the critical role the visual context plays in enriching and disambiguating textual information while grounding events. Furthermore, the fine-tuned Qwen2.5-VL-7B further boosted performance, reaching an F1-score of $65.4\%$ and $65.0\%$ on ED and EAE, respectively. These results demonstrate that lightweight adaptation of a model with only $7$B parameters can still match or surpass the few-shot performance of a $72$B model, reinforcing the effectiveness of parameter-efficient fine-tuning approaches.

\subsection{Error Analysis}

Despite competitive quantitative results, a detailed analysis uncovers consistent shortcomings within each modality are categorized as follows:

\paragraph{Textual EE.} In the textual setting, model errors primarily arise from insufficient \textit{fine-grained understanding}. where models frequently confuse semantically similar event types (e.g., \texttt{Contact:Meet} and \texttt{Contact:Phone-Write}) and struggle with precise argument boundaries, leading to overextension of spans (e.g., predicting ``\textit{Iraqi government forces}'' instead of ``\textit{forces}''). They also generate spurious arguments by misinterpreting context (e.g., labeling the country ``\textit{Uganda}'' as a place argument for a specific attack) or failure to resolve coreference resolution (e.g., identifying pronoun ``\textit{he}'' as an argument). Conversely, critical participants are often overlooked, such as omitting ``\textit{teenager}'' or ``\textit{victim}'' in event descriptions, indicating a potential over-reliance on explicit lexical cues.

\begin{figure}[t]
    \centering
        \includegraphics[width=\linewidth]{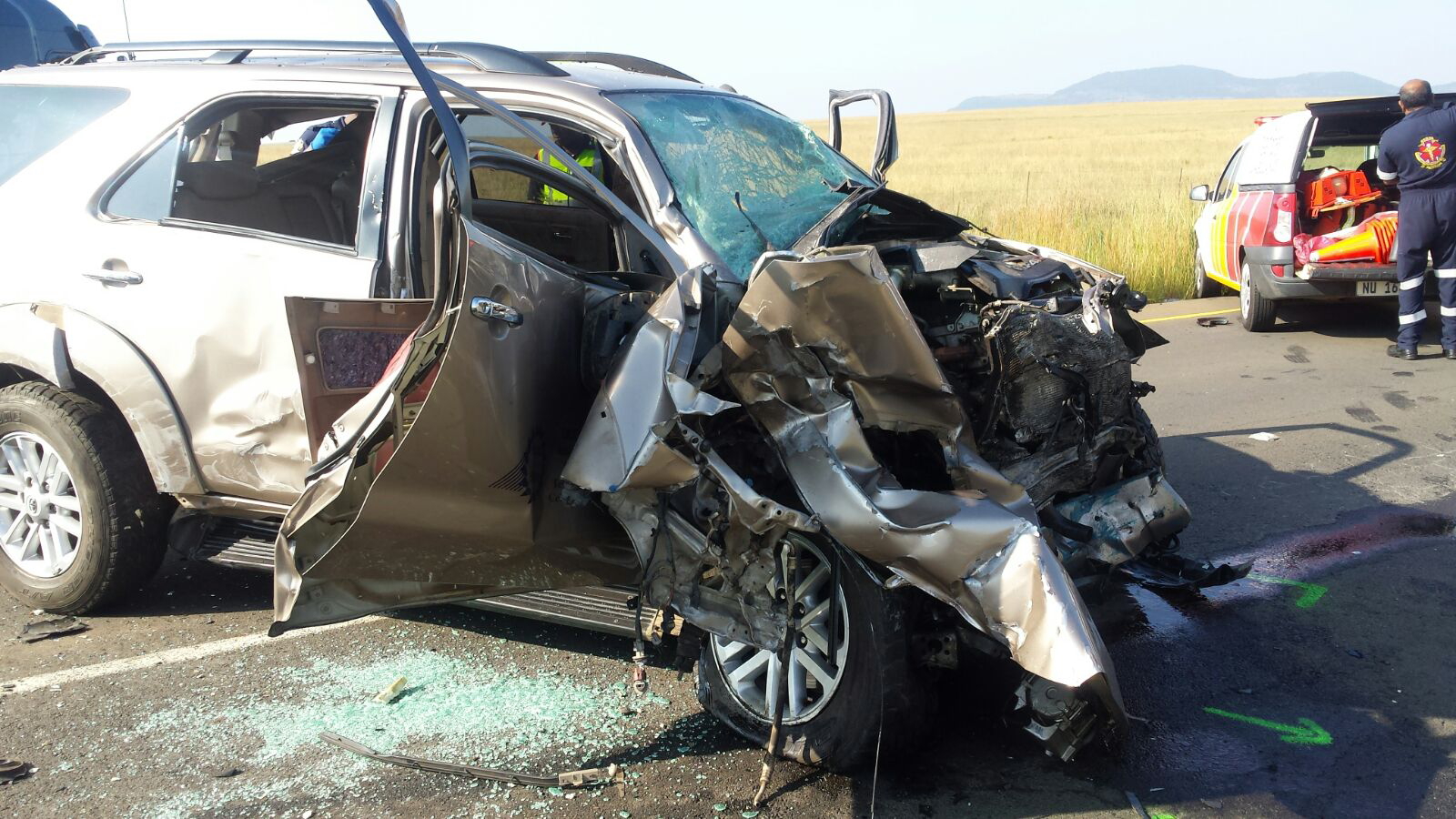}
        \caption{Visual event detection error: \newline Ground Truth: \texttt{Conflict.Attack}, \newline Qwen2.5-VL-72B: \texttt{Movement.Transport}, \newline DeepSeek-VL2: \texttt{Conflict.Demonstrate}.}
        \label{fig:ed-error}
        \vspace{-4mm}
\end{figure}

\begin{figure}
    \centering
        \includegraphics[width=\linewidth]{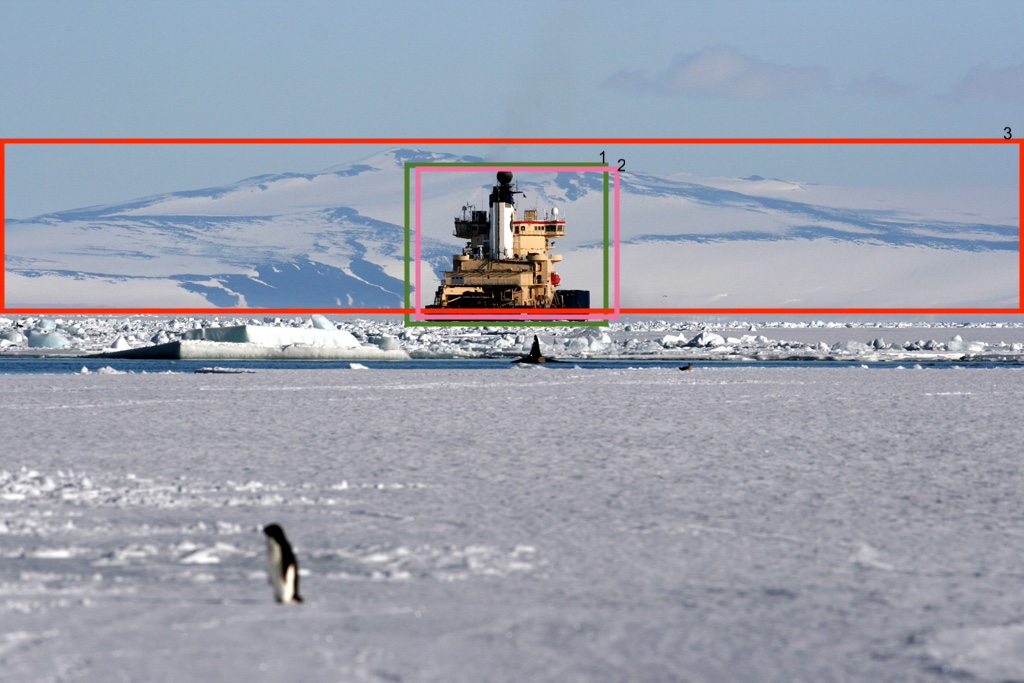}
        \caption{Visual event argument extraction error: \newline
        Ground Truth: \textcolor{PineGreen}{\textbf{Green}} Box,\newline
        Correct Prediction: \textcolor{CarnationPink}{\textbf{Pink}} Box,\newline
        Additional Predicted: \textcolor{BrickRed}{\textbf{Red}} Box. }
        \label{fig:eae_error}
        \vspace{-4mm}
\end{figure}

\paragraph{Visual EE.} In the visual setting, models closely resemble those observed in text. However, they still face difficulties in fine-grained event type classification, as exemplified in Figure~\ref{fig:ed-error}, where an image of a vehicle incident labeled with \texttt{Conflict.Attack} may be misinterpreted by different models. Qwen2.5-VL-72B classifies it as \texttt{Movement.Transport}, while DeepSeek-VL2 predicts \texttt{Conflict.Demonstrate}. Moreover, models often identify visually prominent objects that are not actual event arguments (Figure \ref{fig:eae_error}), resulting in low precision under strict evaluation criteria. Except for the Qwen2.5-VL models benefit from real-world coordinate-aware pretraining \cite{bai2025qwen25vl}, most variants fail to localize objects exactly.

\paragraph{Multimedia EE.} Finally, in the multimedia setting, cross-modal grounding remains a challenge. Notably, errors arising from unimodal processing are not always resolved. For instance, fine-grained semantic confusion between event types or imprecise bounding box localization persist even when complementary information from the other modality is accessible. Furthermore, the internal safety filters of all LVLMs often hinder processing, refusing to process content related to sensitive topics like violence, which directly impacts recall for event types such as \texttt{Conflict:Attack} and \texttt{Life:Die}. 

\section{Conclusion}
We present the first comprehensive evaluation of LVLMs for M²E², assessing their performance in both few-shot and fine-tuning settings. Our results reveal a critical interplay between modality and training strategy: Few-shot LVLMs excel on visual tasks but falter in the texts, a gap that is dramatically closed by fine-tuning. Ultimately, our work confirms that LVLMs excel in combining text and vision modalities, creating powerful cross-modal synergy that yields the best performance, highlighting the significant potential of adapted LVLMs for complex multimedia understanding. We further highlight the persistent challenges in our error analysis, including semantic precision, localization, and cross-modal grounding, which we will design more robust architectures to resolve in future research.

\section*{Limitations}
We acknowledge the following limitations of this study that provide avenues for future work. First, due to the only dataset available, we only experimented in the news domain using the M²E² and ACE 2005 datasets. Second, while we evaluated several representative open-source LVLMs, we did not include closed-source models due to the high cost required, but it does not diminish the contribution and findings of this study.

\section*{Acknowledgements}

We sincerely thank anonymous reviewers for their valuable comments. This research is supported by the Research Development Funding (RDF) (RDF-21-02-044) and Collaborative Research Project (RDS10120240248) at Xi’an Jiaotong-Liverpool University.

\bibliography{custom}
\appendix

\newpage

\onecolumn

\section{Prompt}

\begin{table*}[h]
\centering
\small
\begin{tabular}{lp{12cm}}
\toprule
\textbf{Component} & \textbf{Example for Text-only Event Detection (ED)} \\
\midrule

\textbf{Instruction} & \texttt{Please identify the trigger word and classify the events in the query into one or more appropriate categories based on the given choices. Then, output the trigger word, its position interval, and the option.} \\
\addlinespace 

\textbf{Demonstrations} & 
\texttt{--- Example 1 (Positive) ---} \newline
\texttt{Sentence: Two Chicago police officers take a man into custody during a protest march.} \newline
\texttt{Answer: custody; [8,9]; (D)} \newline
\textit{<Other demonstrations (positive and negative) are provided in the actual prompt.>} \\
\addlinespace

\textbf{Query} & 
\texttt{\{sentence\}} \\
\addlinespace

\textbf{Options} &
\texttt{(A) Movement.Transport (B) Conflict.Attack (C) Conflict.Demonstrate (D) Justice.ArrestJail (E) Contact.PhoneWrite (F) Contact.Meet (G) Life.Die (H) Transaction.TransferMoney} \\
\addlinespace

\textbf{Answer} &
\textit{<placeholder>}
\\

\bottomrule
\end{tabular}
\caption{Example of the few-shot prompt structure for the textual event detection (ED) subtask. The prompt includes a detailed instruction, contrastive demonstrations, and the final query for the model to complete. The structure is adapted for other subtasks, such as argument extraction.}
\label{tab:prompt_example}
\end{table*}

\newpage

\section{Full Experimental Results}
\label{sec:full}

\begin{table*}[htbp]
  \centering
  \small
  \setlength{\tabcolsep}{12pt} 
  \begin{tabular}{l|l|rrr|rrr}
    \toprule
    \multicolumn{2}{c|}{\multirow{2}{*}{Model}} & \multicolumn{3}{c|}{Event Mention} & \multicolumn{3}{c}{Argument Role} \\
    \cmidrule(lr){3-5} \cmidrule(lr){6-8}
    \multicolumn{2}{c|}{} & P & R & F1 & P & R & F1 \\
    \midrule
    \multirow{15}{*}{\rotatebox[origin=c]{90}{Textual}}
    & JMEE & $42.5$ & $58.2$ & $48.7$ & $22.9$ & $28.3$ & $25.3$ \\
    & GAIL & $43.4$ & $53.5$ & $47.9$ & $23.6$ & $29.2$ & $26.1$ \\
    & WASE-T & $42.3$ & $58.4$ & $48.2$ & $21.4$ & $30.1$ & $24.9$ \\
    & WASE$_{att}$ & $37.6$ & $66.8$ & $48.1$ & $27.5$ & $33.2$ & $30.1$ \\
    & WASE$_{obj}$ & $42.8$ & $61.9$ & $50.6$ & $23.5$ & $30.3$ & $26.4$ \\
    & UniCL & $49.1$ & $59.2$ & $53.7$ & $27.8$ & $34.3$ & $30.7$ \\
    & CAMEL & $45.1$ & $71.8$ & $55.4$ & $24.8$ & $41.8$ & $31.1$ \\
    & X-MTL & $49.7$ & $65.7$ & $\mathbf{56.6}$ & $34.6$ & $37.6$ & $36.0$ \\
    & Qwen2-VL-7B & $13.3$ & $19.7$ & $15.9$ & $75.0$ & $2.6$ & $5.1$ \\
      & Qwen2.5-VL-7B & $70.0$ & $3.7$ & $6.9$ & $4.2$ & $1.4$ & $2.1$ \\
    & DeepSeek-VL2-4.5B & $30.5$ & $12.0$ & $17.5$ & $5.6$ & $2.2$ & $3.1$ \\
    & Qwen2.5-VL-72B & $80.0$ & $12.0$ & $20.6$ & $10.1$ & $7.5$ & $8.6$ \\
    & Qwen2-VL-7B-LoRA & $46.8$ & $59.6$ & $52.5$ & $16.3$ & $24.1$ & $19.5$ \\
    & Qwen2.5-VL-7B-LoRA & $41.5$ & $52.6$ & $46.4$ & $35.5$ & $50.3$ & $\mathbf{41.6}$ \\
    \midrule
    \multirow{14}{*}{\rotatebox[origin=c]{90}{Visual}}
    & CLIP-Event & $41.3$ & $72.8$ & $52.7$ & $21.1$ & $13.1$ & $17.1$ \\
    & WASE-$V_{att}$ & $29.7$ & $61.9$ & $40.1$ & $9.1$ & $10.2$ & $9.6$ \\
    & WASE-$V_{obj}$ & $28.6$ & $59.2$ & $38.7$ & $13.3$ & $9.8$ & $11.2$ \\
    & WASE$_{att}$ & $32.3$ & $63.4$ & $42.8$ & $9.7$ & $11.1$ & $10.3$ \\
    & WASE$_{obj}$ & $43.1$ & $59.2$ & $49.9$ & $14.5$ & $10.1$ & $11.9$ \\
    & UniCL & $54.6$ & $60.9$ & $57.6$ & $16.9$ & $13.8$ & $15.2$ \\
    & CAMEL & $52.1$ & $66.8$ & $58.5$ & $21.4$ & $28.4$ & $24.4$ \\
    & X-MTL & $73.1$ & $70.3$ & $71.7$ & $33.2$ & $31.3$ & $32.2$ \\
    & Qwen2-VL-7B & $69.6$ & $62.1$ & $65.6$ & $3.3$ & $4.3$ & $3.8$ \\
    & Qwen2.5-VL-7B & $54.3$ & $47.3$ & $50.5$ & $23.7$ & $35.3$ & $28.3$ \\
    & DeepSeek-VL2-4.5B & $39.3$ & $76.4$ & $51.9$ & $0.7$ & $0.7$ & $0.7$ \\
    & Qwen2.5-VL-72B & $73.0$ & $70.6$ & $\mathbf{71.8}$ & $62.8$ & $71.9$ & $\mathbf{67.0}$ \\
    & Qwen2-VL-7B-LoRA & $49.7$ & $74.4$ & $59.6$ & $23.5$ & $4.3$ & $7.3$ \\
    & Qwen2.5-VL-7B-LoRA & $48.5$ & $72.5$ & $58.1$ & $10.8$ & $19.6$ & $14.0$ \\
    \midrule
    \multirow{11}{*}{\rotatebox[origin=c]{90}{Multimedia}}
    & WASE$_{att}$ & $38.2$ & $67.1$ & $49.1$ & $18.6$ & $21.6$ & $19.9$ \\
    & WASE$_{obj}$ & $43.0$ & $62.1$ & $50.8$ & $19.5$ & $18.9$ & $19.2$ \\
    & UniCL & $44.1$ & $67.7$ & $53.4$ & $24.3$ & $22.6$ & $23.4$ \\
    & CAMEL & $55.6$ & $59.5$ & $57.5$ & $31.4$ & $35.1$ & $33.2$ \\
    & X-MTL & $78.3$ & $57.3$ & $66.2$ & $40.3$ & $42.6$ & $41.4$ \\
    & Qwen2-VL-7B & $65.4$ & $53.7$ & $60.0$ & $20.6$ & $21.9$ & $21.2$ \\
    & Qwen2.5-VL-7B & $62.9$ & $37.2$ & $46.8$ & $23.9$ & $14.9$ & $18.4$ \\
    & DeepSeek-VL2-4.5B & $41.3$ & $79.6$ & $54.4$ & $8.1$ & $5.8$ & $6.8$ \\
    & Qwen2.5-VL-72B & $84.0$ & $81.4$ & $\mathbf{82.5}$ & $56.1$ & $71.9$ & $63.0$ \\
    & Qwen2-VL-7B-LoRA & $41.7$ & $79.3$ & $54.7$ & $55.3$ & $70.8$ & $62.1$ \\
    & Qwen2.5-VL-7B-LoRA & $56.6$ & $77.4$ & $65.4$ & $54.7$ & $80.1$ & $\mathbf{65.0}$ \\
    \bottomrule
  \end{tabular}
    \caption{Full results on the M²E² dataset, including all baseline models for comparison, which is the unabridged version of the condensed table presented in the main paper. \textbf{Bold} indicates the best result in each section.}
\label{tab:full_results}
\end{table*}

\end{document}